\documentclass[11pt,letterpaper]{article}
\usepackage[letterpaper]{geometry}
\usepackage{acl2012}
\usepackage{fixltx2e}
\usepackage{times}
\usepackage{verbatim}
\usepackage{latexsym}
\usepackage{amsmath}
\usepackage{amsfonts}
\usepackage{multirow}
\usepackage[caption=false]{subfig}
\usepackage{graphicx}
\usepackage[table,x11names]{xcolor}
\usepackage{tikz,pgfplots}
\usepackage{url}
\makeatletter
\newcommand{\@BIBLABEL}{\@emptybiblabel}
\newcommand{\@emptybiblabel}[1]{}
\makeatother
\usepackage[hidelinks]{hyperref}

\title{Sparse Coding of Neural Word Embeddings for Multilingual Sequence Labeling}

\author{G\'{a}bor Berend\\
  Department of Informatics\\
  University of Szeged \\
  2~\'{A}rp\'{a}d t\'{e}r, 6720 Szeged, Hungary \\
  {\tt berendg@inf.u-szeged.hu}
\\}

\date{}

\begin{document}
\maketitle
\begin{abstract}
In this paper we propose and carefully evaluate a sequence labeling framework which solely utilizes sparse indicator features derived from dense distributed word representations. The proposed model obtains (near) state-of-the art performance for both part-of-speech tagging and named entity recognition for a variety of languages. Our model relies only on a few thousand sparse coding-derived features, without applying any modification of the word representations employed for the different tasks. The proposed model has favorable generalization properties as it retains over 89.8\% of its average POS tagging accuracy when trained at 1.2\% of the total available training data, i.e.~150 sentences per language. 
\end{abstract}

\section{Introduction}

Determining the linguistic structure of natural language texts based on rich hand-crafted features has a long-going history in natural language processing. The focus of traditional approaches has mostly been on building linguistic analyzers for a \textit{particular kind of analysis}, which often leads to the incorporation of extensive linguistic and/or domain knowledge for defining the feature space. Consequently, traditional models easily become language and/or task specific resulting in improper generalization properties.

A new research direction has emerged recently, which aims at building more general models that require far less feature engineering or none at all. These advancements of natural language processing, pioneered by \newcite{Bengio:2003:NPL:944919.944966}, followed by \newcite{Collobert:2008:UAN:1390156.1390177}, \newcite{Collobert:2011:NLP:1953048.2078186}, \newcite{DBLP:journals/corr/abs-1301-3781} among others, employ a different philosophy. The objective of these works is to find representations for linguistic phenomena in an unsupervised manner by relying on large amounts of text.

Natural language phenomena are extremely sparse by their nature, whereas continuous word embeddings employ dense representations of words. In our paper we empirically verify via rigorous experiments that turning these dense representations into a much sparser (yet denser than one-hot encoding) form can help in keeping the most salient parts of word representations that are highly suitable for sequence models.

Furthermore, our experiments reveal that our proposed model performs substantially better than traditional feature-rich models in the absence of abundant training data. Our proposed model also has the advantage of performing well on multiple sequence labeling tasks without any modification in the applied word representations thanks to the sparse features derived from continuous word representations.

Our work aims at introducing a novel sequence labeling model solely utilizing features derived from the sparse coding of continuous word embeddings. Even though sparse coding has been utilized in NLP prior to us \cite{faruqui-EtAl:2015:ACL-IJCNLP,chen-EtAl:2016:P16-11}, to the best of our knowledge, we are the first to propose a sequence labeling framework incorporating it with the following contributions:
\begin{itemize}
\item we show that the proposed sparse representation is general as sequence labeling models trained on them achieve (near) state-of-the-art performances for both POS tagging and NER,
\item we show that the representation is general in the other sense, that it produces reasonable results for more than 40 treebanks for POS tagging,
\item we rigorously compare different sparse coding approaches in conjunction with differently trained continuous word embeddings,
\item we highlight the favorable generalization properties of our model in settings when access to a very limited training corpus is assumed,
\item we release the sparse word representations determined for our experiments at \url{https://begab.github.io/sparse_embeds} to ensure the replicability of our results and to foster further multilingual NLP research.
\end{itemize}

\section{Related work}

The line of research introduced in this paper relies on distributed word representations \cite{polyglot:2013:ACL-CoNLL} and dictionary learning for sparse coding \cite{Mairal:2010:OLM:1756006.1756008} and also shows close resemblance to~\cite{faruqui-EtAl:2015:ACL-IJCNLP}.

\subsection{Distributed word representations}
Distributed word representations assign some relatively low-dimensional, dense vectors to each word in a corpus such that words with similar context and meaning tend to have similar representations.
From an algebraic point of view, the embedding of word $i$ having index $idx_i$ in a vocabulary $V$ can be thought of as the result of a matrix-vector multiplication $W \mathbf{1}_{i},$ where the $i^{th}$ column of matrix $W \in \mathbb{R}^{k \times |V|}$ contains the $k$-dimensional ($k \ll |V|$) embedding for word $i$ and vector $\mathbf{1}_{i}\in \mathbb{R^{|V|}}$ is the one-hot representation of word $i$. The one-hot representation of word $i$ is such a vector, which contains zeros for all of its entries except for index $idx_i$ where it stores a one. Depending on how the columns of $W$ (i.e.~the word embeddings) get determined, we could distinguish a plethora of approaches \cite{Bengio:2003:NPL:944919.944966,DBLP:journals/corr/LebretL13,mnih-nips-2013,Collobert:2008:UAN:1390156.1390177,DBLP:journals/corr/abs-1301-3781,pennington2014glove}.

Prediction-based distributed word embedding approaches such as \texttt{word2vec}~\cite{DBLP:journals/corr/abs-1301-3781} have been conjectured to have superior performance over count-based word representations \cite{baroni-dinu-kruszewski:2014:P14-1}.
However, as \newcite{DBLP:journals/corr/LebretC14}, \newcite{DBLP:journals/tacl/LevyGD15} and \newcite{qu-EtAl:2015:CoNLL} point out count-based distributional models can perform on par with prediction-based distributed word embedding models. \newcite{DBLP:journals/tacl/LevyGD15} illustrate that the effectiveness of neural word embeddings largely depend on the selection of model hyperparameters and other design choices.

According to these findings, in order to avoid any hassles of tuning the hyperparameters of the word embedding model employed, we rather primarily use the publicly available pre-trained \texttt{polyglot} word embeddings of \cite{polyglot:2013:ACL-CoNLL} without any task specific modification for our experiments. A key thing to note is that \texttt{polyglot} word embeddings are not tailored for any specific language analysis task such as POS tagging or NER. These word embeddings are instead trained in a manner favoring the word analogy task introduced by \newcite{DBLP:journals/corr/MikolovSCCD13}. The \texttt{polyglot} project distributes word embeddings for more than 100 languages. \newcite{polyglot:2013:ACL-CoNLL} also report results on POS tagging, however, word representations they apply for these experiments are different from the task-agnostic representations they made publicly available.

There have been further previous research conducted on training neural networks for learning distributed word representations for various specific language analysis tasks. \newcite{Collobert:2011:NLP:1953048.2078186} propose neural network architectures to four natural language processing tasks, i.e.~POS tagging, named entity recognition, semantic role labeling and chunking. \newcite{Collobert:2011:NLP:1953048.2078186} train word representations on large amounts of unannotated texts from Wikipedia, then update the pre-trained word representations for the individual tasks. Our approach is different in that we do not update our word representations for the different tasks and most importantly that we use successfully the features derived from sparse coding in a log-linear model instead of a neural network architecture. A final difference to \cite{Collobert:2011:NLP:1953048.2078186} is that we experiment with a much wider range of languages while they report results for English only.

\newcite{qu-EtAl:2015:CoNLL} evaluate the impacts of choosing different embedding methods on four sequence labeling tasks, i.e.~POS tagging, NER, syntactic chunking and multiword expression identification. The hand-crafted features they employ for POS tagging and NER are the same as in \newcite{Collobert:2011:NLP:1953048.2078186} and \newcite{Turian:2010:WRS:1858681.1858721}. 

\subsection{Sparse coding}
The general goal of sparse coding is to express signals in the form of \textit{sparse} linear combination of basis vectors and the task of finding an appropriate set of basis vectors is referred to as the dictionary learning problem \cite{Mairal:2010:OLM:1756006.1756008}. Generally, given a data matrix $X\in \mathbb{R}^{k \times n}$ with its $i^{th}$ column $\mathbf{x}_i$ representing the $i^{th}$ $k$-dimensional signal, the task is to find $D\in \mathbb{R}^{k \times m}$ and $\alpha \in \mathbb{R}^{m \times n}$, such that $X \approx D \alpha$. This can be formalized into an $\ell_1$-regularized linear least-squares minimization problem having the form
\begin{equation}
\min\limits_{D \in \mathcal{C}, \alpha} \frac{1}{2n} \sum_{i=1}^{n} \left(\lVert \mathbf{x}_i-D \boldsymbol{\alpha}_i \rVert_2^2 + \lambda \lVert \boldsymbol{\alpha}_i \rVert_1 \right),
\label{SPAMS_objective}
\end{equation}
with $\mathcal{C}$ being the convex set of matrices that comprise of column vectors having an $\ell_2$ norm at most one, matrix $D$ acts as the shared dictionary across the signals, and the columns of the sparse matrix $\alpha$ contains the coefficients for the linear combinations of each of the $n$ observed signals.

Performing sparse coding of word embeddings has recently been proposed by \newcite{faruqui-EtAl:2015:ACL-IJCNLP}, however, the objective function they optimize differs from (\ref{SPAMS_objective}). In Section~\ref{sec:experiments}, we compare the effects of employing different sparse coding paradigms including the ones in \cite{faruqui-EtAl:2015:ACL-IJCNLP}.

In their work, \newcite{yoga:2015:hierarchical} proposed an efficient learning algorithm for determining hierarchically organized sparse word representations using stochastic proximal methods. Most recently, \newcite{Sun:Sparse} have proposed an online learning algorithm using regularized dual averaging to directly obtain $\ell_1$ regularized continuous bag of words (CBOW) representations \cite{DBLP:journals/corr/abs-1301-3781} without the need to determine dense CBOW representations first.

\section{Sequence labeling framework}
\label{sec:framework}

This section introduces the sequence labeling framework we use for both POS tagging and NER. Since our goal is to measure the effectiveness of sparse word embeddings alone, we do not apply any features based on gazetters, capitalization patterns or character suffixes.

As described previously, word embedding methods turn high-dimensional (i.e.~as many dimensional as many words are in the vocabulary) and extremely sparse (i.e.~containing only one non-zero element at the vocabulary index of the word it represents) one-hot encoded representation of words into a dense embedding of much lower dimensionality $k$.

In our work, instead of taking the low dimensional dense word embeddings, we use a dictionary learning approach to obtain sparse codings for the embedded word representations. Formally, given the lookup matrix $W \in \mathbb{R}^{k \times |V|}$ which contains the embedding vectors, we learned $D \in \mathbb{R}^{k \times m}$ being the dictionary matrix shared across all the embedding vectors and $\alpha \in \mathbb{R}^{m \times |V|}$ containing sparse linear combination coefficients for each of the word embeddings in such a way that $\lVert W-D\alpha\rVert_F^2 + \lambda \lVert \alpha\rVert_1$ is minimized.

Once the dictionary matrix $D$ is learned, the sparse linear combination coefficients $\boldsymbol{\alpha}_i$ can easily be determined for a word embedding vector $\mathbf{w}_i$ by solving an $\ell_1$-regularized linear least-squares minimization problem \cite{Mairal:2010:OLM:1756006.1756008}. We define features based on vector $\boldsymbol{\alpha}_i$ by taking the signs and indices of its non-zero coefficients, that is
\begin{equation}
f(\mathbf{w}_i)=\{sign(\boldsymbol{\alpha}_i[j]) j \mid \boldsymbol{\alpha}_i[j] \neq 0\},
\label{sparse_feature}
\end{equation}
$\boldsymbol{\alpha}_i[j]$ denoting the $j^{th}$ coefficient in the sparse vector $\boldsymbol{\alpha}_i$.
The intuition behind this feature is that words with similar meaning are expected to use an overlapping set of basis vectors from dictionary $D$. Incorporating the signs of coefficients into the feature function can help to distinguish cases when a basis vector takes part in the reconstruction of a word representation 'destructively' or 'constructively'.

When assigning features to a target word at some position within a sentence, we determine the same set of feature functions for the target word itself and its neighboring words of window size 1. Experiments with window size 2 were also performed, however, these results get omitted for brevity as they do not substantially differ from the ones obtained when a window size of 1 is applied.

We then use the previously described set of features in a linear chain CRF~\cite{Lafferty:2001:CRF:645530.655813} using CRFsuite~\cite{CRFsuite}. As our goal is not to tweak the proposed framework by extreme hyperparameter tuning, we simply use the default settings of CRFsuite to learn model parameters. That is, the coefficients for $\ell_1$ and $\ell_2$ regularization is set to $1.0$ and $0.001$, respectively.

\section{Experiments}
\label{sec:experiments}

We primarily rely on the SPArse Modeling Software\footnote{\url{http://spams-devel.gforge.inria.fr/}} (SPAMS) \cite{Mairal:2010:OLM:1756006.1756008} for performing sparse coding of distributed word representations. For dictionary learning as formulated in Equation~\ref{SPAMS_objective} one should choose $m$ and $\lambda$, controlling the number of the basis vectors and the regularization coefficient affecting the sparsity of $\alpha$, respectively. Starting with $m=256$ and doubling it at each iteration, our preliminary investigations showed a steady growth in the usefulness of sparse word representations as a function of $m$, plateauing at $1024$, for which reason we set $m$ to that value for further experiments.

\subsection{Baseline methods}

\paragraph{Brown clustering}

Various studies have identified Brown clustering~\cite{Brown:1992:CNG:176313.176316} as a useful source of feature generation for sequence labeling tasks \cite{Ratinov:2009:DCM:1596374.1596399,Turian:2010:WRS:1858681.1858721,owoputi-EtAl:2013:NAACL-HLT,stratos-collins:2015:VSM-NLP,derczynski-chester-bogh:2015:RANLP2015}. We should note that sparse coding can also be viewed as a kind of clustering which -- unlike Brown clustering -- has the capability of assigning word forms to multiple clusters at a time (corresponding the non-zero coefficients in $\alpha$).

We thus define a linear chain CRF relying on features from the Brown cluster identifier of words as one of our baseline approach. Since Brown clustering defines a hierarchical clustering over words, cluster supersets can easily function as features. We employ the frequently used approach of generating features from the length-$p$ ($p\in\{4,6,10,20\}$) prefixes of Brown cluster identifiers similar to \newcite{Ratinov:2009:DCM:1596374.1596399} and \newcite{Turian:2010:WRS:1858681.1858721}.

In our experiments we use the implementation by \newcite{liang05meng} for performing Brown clustering\footnote{\url{https://github.com/percyliang/brown-cluster}}. We provide the very same Wikipedia articles as input text for determining Brown clusters that are used for training the \texttt{polyglot}\footnote{\url{https://sites.google.com/site/rmyeid/projects/polyglot}} word embeddings. We also set the number of Brown clusters to be identified to 1024, that is the number of basis vectors applied during sparse coding (cf.~$D\in\mathbb{R}^{64 \times 1024}$).

\paragraph{Feature-rich representation}

We report results relying on linear chain CRFs that assign standard state-of-the-art feature-rich representation to sequences.
We apply the very same features and feature templates included in the POS tagging model of CRFSuite\footnote{\url{http://github.com/chokkan/crfsuite/blob/master/example/pos.py}}.
We summarize these features in Table~\ref{feature_templates}, where $\oplus$ denotes the binary operator which defines features as a combination of word forms at different (not necessarily contiguous) positions of a sentence.

\begin{table}
\small
\centering
\begin{tabular}{l|l|lr}
\#&Level  & Feature name &  \\ \hline
1& char & $isNumber(w_t)$ & \\
2& char & $isTitleCase(w_t)$ & \\
3&char & $isNonAlnum(w_t)$ & \\
4&char & $prefix(w_t, i)$ & $ 1 \le i \le 4$ \\
5&char & $suffix(w_t, i)$ & $ 1 \le i \le 4$ \\ \hline
6&word & $w_{t+j}$ & $-2 \le j \le 2$ \\
7&word & $w_{t} \oplus w_{t+i}$ & $1 \le i \le 9$  \\
8&word & $w_{t} \oplus w_{t-i}$ & $1 \le i \le 9$  \\
9&word & $\oplus_{i=t+j}^{t+j+1}w_i$ & $ -2 \le j \le 1$  \\
10&word & $\oplus_{i=t+j}^{t+j+2}w_i$ & $ -2 \le j \le 0$  \\
11&word & $\oplus_{i=t+j-1}^{t+j+2}w_i$ & $ -1 \le j \le 0$  \\
12&word & $\oplus_{i=t-2}^{t+2}w_i$ &  \\
\end{tabular}
\caption{Features and feature templates applied by our feature-rich baseline for target word $w_t$. $\oplus$ is a binary operator forming a feature from words and their relative positions by combining them together.}
\label{feature_templates}
\end{table}

We use the same pool of features described in Table~\ref{feature_templates} for both POS tagging and NER. The reason why we do not adjust the feature-rich representation employed as our baseline for the different tasks is that we do not alter our representation in any way when using our sparse coding-based model either.

Note that features numbered up to 5 in Table~\ref{feature_templates} operate at the character-level, whereas our proposed framework solely uses features derived from the sparse coding of word forms. We thus distinguish two feature-rich baselines, i.e.~FR\textsubscript{w+c} including both word and character-level features and FR\textsubscript{w} which treats word forms as atomic units to define features.

\paragraph{Using dense word representations}
As our ultimate goal is to demonstrate the usefulness of sparse features derived from dense word representations, it is important to address the question whether sparse word representations are more beneficial for sequence labeling tasks compared to their dense counterparts. For this end, we came up with a similar model to the one proposed in Section~\ref{sec:framework} except for the fact that we used the original dense word representations for inducing features.

According to this modification, we made the following change in our feature function: instead of calculating Equation~(\ref{sparse_feature}) for some word $i$, the modified feature function we use for this baseline is
\begin{equation*}
f(\mathbf{w}_i)=\{j:\mathbf{w}_i[j] \mid \forall j \in \{1,\ldots,k\}\}.
\label{dense_feature}
\end{equation*}
That is, instead of relying on the nonzero values in $\boldsymbol{\alpha}_i$, each word is characterized by its $k$ real-valued coordinates in the embedding space. In order to notationally distinguish sparse and dense representations, we add subscript SC when we refer to a sparse coded version of some word embedding (e.g.~SG\textsubscript{SC}).

\subsection{POS tagging experiments}

Even though it is reasonable to assume that languages share a common coarse set of linguistic categories, linguistic resources used to have their own notations for part-of-speech tags. The first notable attempt trying to canonize the multiple tag sets existing is the Google universal part-of-speech tags introduced by \newcite{PETROV12.274} arguing that the POS tags of various tagging schemes can be mapped to 12 language-independent part-of-speech tags.

There is a recent initiative of universal dependencies (UD) \cite{DBLP:conf/cicling/Nivre15}, which aims at providing a unified notation for multiple linguistic phenomena, including part-of-speech tags as well. The POS tag set proposed for UD has 17 partially overlapping categories to the ones defined by~\newcite{PETROV12.274}.

\subsubsection{Experiments using CoNLL 2006/07 data}
\label{sec:CoNLLX}

\begin{table}
\small
\centering
\begin{tabular}{l|l}
 Language & Source \\ \hline
 bg & BTB/CoNLL06 \shortcite{bg} \\
 da & DDT/CoNLL06 \shortcite{da} \\
 de & Tiger/CoNLL06 \shortcite{de} \\
 en & Penn Treebank \shortcite{penn} \\
 es & Cast3LB/CoNLL06 \shortcite{es} \\
 hu & Szeged Treebank/CoNLL07 \shortcite{hu} \\
 it & ISST/CoNLL07 \shortcite{it} \\
 nl & Alpino/CoNLL06 \shortcite{nl} \\
 pt & Floresta Sintá(c)tica/CoNLL06 \shortcite{pt} \\
 sl & SDT/CoNLL06 \shortcite{sl} \\
 sv & Talbanken05/CoNLL06 \shortcite{sv} \\
 tr & METU-Sabanci/CoNLL07 \shortcite{tr} \\
\end{tabular}
\caption{Treebanks used for POS tagging experiments from the CoNLL 2006/07 shared task.}
\label{CoNLL_languages}
\end{table}

We use 12 treebanks in the CoNLL-X format from the CoNLL-2006/07 \cite{Buchholz:2006:CST:1596276.1596305,nivre-EtAl:2007:EMNLP-CoNLL2007} shared tasks. The complete list of the treebanks included in our experiments is presented in Table~\ref{CoNLL_languages}.

We rely on the official scripts released by \newcite{PETROV12.274}\footnote{\url{https://github.com/slavpetrov/universal-pos-tags}} for mapping the treebank specific POS tags to the Google universal POS tags in order to obtain results comparable across languages.
For our experiments we used the original CoNLL-X train/test splits of the treebanks.

\begin{figure}
\centering
\begin{tikzpicture}[scale=0.7]
    \begin{axis}[
        width  = \columnwidth,
        major x tick style = transparent,ybar,
        ybar=\pgflinewidth,
        bar width=4pt,
        ymajorgrids = true,
        yminorgrids = true,
        minor ytick={55,65,75,85,95},
        ylabel = {Coverage(\%)},
        symbolic x coords={bg,da,de,en,es,hu,it,nl,pt,sl,sv,tr,Avg.},
        xtick = data,
        scaled y ticks = false,
        enlarge x limits=0.03,
        xticklabel style={text height=1ex, text width=1ex},
        legend cell align=left, legend columns=2,
        legend style={draw=none,at={(1,1)}},
        ylabel style={at={(0.11,0.7)}, anchor=south east},
        cycle list = {red,blue!35}
    ]
        \addplot+[fill,style={mark=none}]
            coordinates {(bg,91.63) (da,93.51) (de,89.79) (en,96.90) (es,93.71) (hu,87.80) (it,92.92) (nl,94.46) (pt,92.56) (sl,92.55) (sv,94.34) (tr,85.69) (Avg.,93.25)};

        \addplot+[fill,style={mark=none}]
             coordinates {(bg,67.22) (da,71.37) (de,49.06) (en, 73.38) (es,75.24) (hu,60.73) (it, 82.41) (nl,68.05) (pt,66.20) (sl, 73.93) (sv,64.99) (tr,58.43) (Avg.,64.32)};
        \legend{Token,Word form}
    \end{axis}
\end{tikzpicture}
\caption{Token and word form-level coverages of the word vectors against the combined train/test sets of the CoNLL-2006/07 POS tagging datasets.}
\label{fig:coverages}
\end{figure}
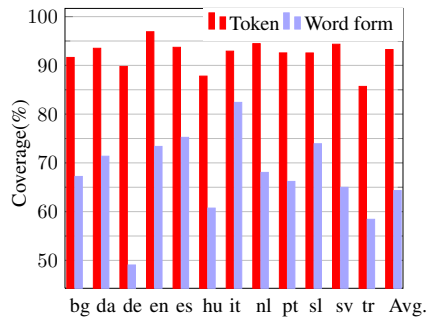

A key factor for the efficiency of our proposed model resides in the coverage of word embeddings, i.e.~the proportion of tokens/word forms with distributed representation determined for. Figure~\ref{fig:coverages} depicts these coverage scores calculated over the merged training and test sets for the different languages. Figure~\ref{fig:coverages} reveals that a substantial amount of tokens has distributed representation defined for (around 90\% for the majority of languages, except for Turkish where it is 5 point less).
Token coverages of the word embeddings are most likely affected by the morphological richness of the languages and the elaborateness of the corresponding Wikipedia used for training word embeddings.

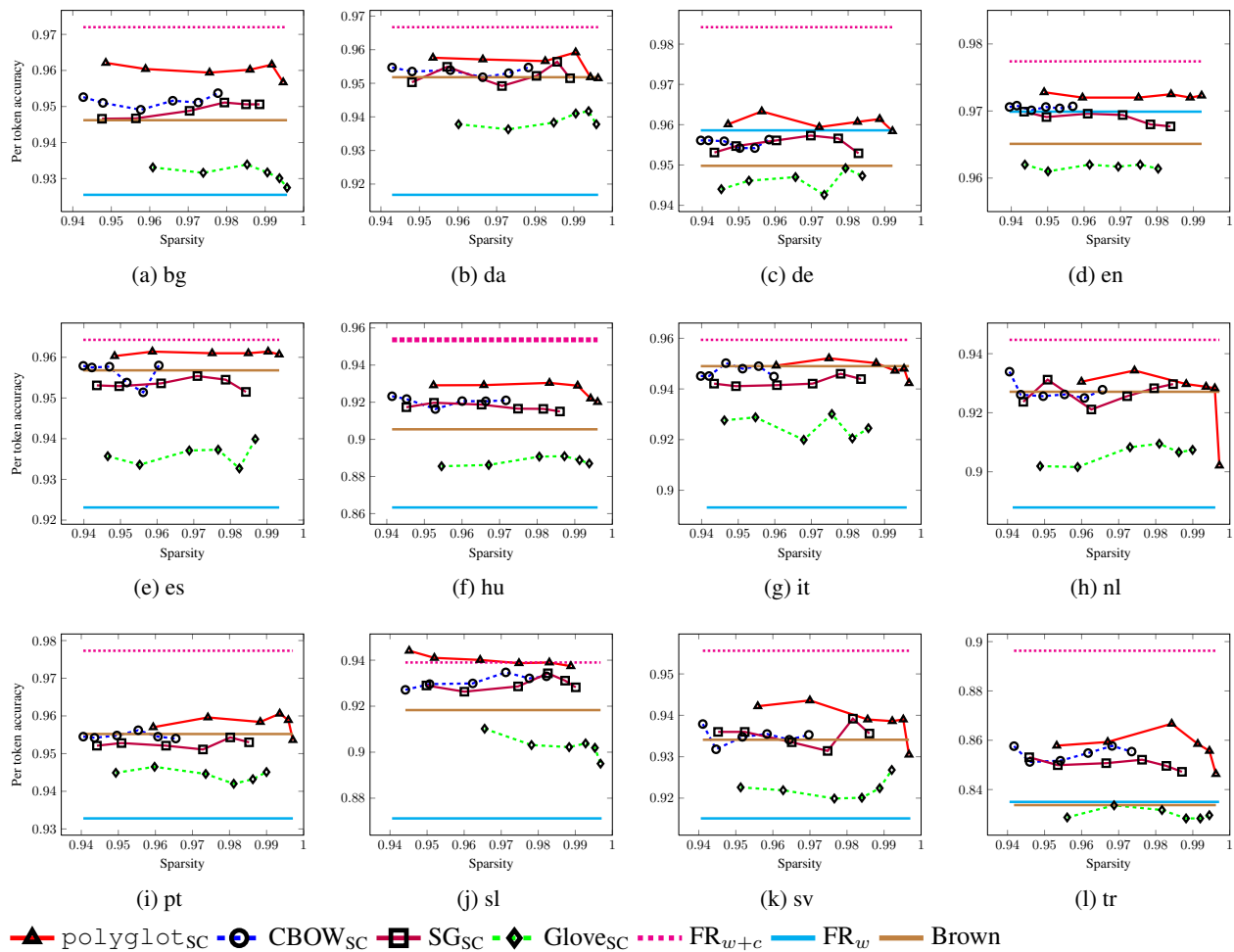
\begin{figure*}[!ht]
	\subfloat[bg]{
	\begin{tikzpicture}[scale=.48]
	\begin{axis}[xmax=1,xlabel=Sparsity,ylabel=Per token accuracy,legend columns=7,legend to name=named,legend entries={\texttt{polyglot}\textsubscript{SC},CBOW\textsubscript{SC},SG\textsubscript{SC},Glove\textsubscript{SC},FR$_{w+c}$,FR$_{w}$,Brown,TnT}, legend style={draw=none,font=\small}]
	 \addplot[line width=1.8pt, mark=*, mark size=3, color=red, mark=triangle, mark options={solid, draw=black}] coordinates {(0.94857,0.96210) (0.95891,0.96040) (0.97554,0.95940) (0.98606,0.96020) (0.99173,0.96160) (0.99469,0.95670)};
	 \addplot[line width=1.8pt, mark=*, mark size=3, color=blue, dashed, mark=o, mark options={solid, draw=black}] coordinates {(0.94273,0.95260) (0.94784,0.95100) (0.95767,0.94910) (0.96598,0.95160) (0.97261,0.95110) (0.97773,0.95370)};
	 \addplot[line width=1.8pt, mark=*, mark size=3, color=purple, mark=square, mark options={solid, draw=black}] coordinates {(0.94758,0.94660) (0.95640,0.94670) (0.97036,0.94880) (0.97941,0.95110) (0.98502,0.95060) (0.98855,0.95060)};
	 \addplot[line width=1.8pt, mark=*, mark size=3, color=green, dashed, mark=diamond, mark options={solid, draw=black}] coordinates {(0.96080,0.93310) (0.97390,0.93160) (0.98530,0.93390) (0.99058,0.93170) (0.99372,0.93010) (0.99569,0.92750)};
	 \addplot[line width=2pt, dotted, color=magenta] coordinates {(0.94273, 0.9720) (0.99569, 0.9720)};
	 \addplot[line width=2pt, color=cyan] coordinates {(0.94273, 0.9255) (0.99569, 0.9255)};
	 \addplot[line width=2pt, color=brown] coordinates{(0.94273, 0.9462) (0.99569, 0.9462)};
	\end{axis}
	\end{tikzpicture}
	}~
	\subfloat[da]{
	\begin{tikzpicture}[scale=.48]
	\begin{axis}[xmax=1,xlabel=Sparsity]
	 \addplot[line width=1.8pt, mark=*, mark size=3, color=red, mark=triangle, mark options={solid, draw=black}] coordinates {(0.95349,0.95760) (0.96642,0.95710) (0.98260,0.95660) (0.99049,0.95920) (0.99432,0.95180) (0.99631,0.95150)};
	 \addplot[line width=1.8pt, mark=*, mark size=3, color=blue, dashed, mark=o, mark options={solid, draw=black}] coordinates {(0.94289,0.95470) (0.94809,0.95350) (0.95799,0.95390) (0.96640,0.95180) (0.97314,0.95300) (0.97824,0.95470)};
	 \addplot[line width=1.8pt, mark=*, mark size=3, color=purple, mark=square, mark options={solid, draw=black}] coordinates {(0.94805,0.95030) (0.95717,0.95490) (0.97138,0.94920) (0.98032,0.95220) (0.98570,0.95640) (0.98904,0.95150)};
	 \addplot[line width=1.8pt, mark=*, mark size=3, color=green, dashed, mark=diamond, mark options={solid, draw=black}] coordinates {(0.96013,0.93780) (0.97302,0.93630) (0.98481,0.93830) (0.99051,0.94100) (0.99384,0.94170) (0.99585,0.93780)};
	 \addplot[line width=2pt, dotted, color=magenta] coordinates {(0.94289, 0.9667) (0.99631, 0.9667)};
	 \addplot[line width=2pt,  color=cyan] coordinates {(0.94289, 0.9168) (0.99631, 0.9168)};
	 \addplot[line width=2pt, color=brown] coordinates{(0.94289, 0.9518) (0.99631, 0.9518)};
	\end{axis}
	\end{tikzpicture}
	}~
	\subfloat[de]{
	\begin{tikzpicture}[scale=.48]
	\begin{axis}[xmax=1,xlabel=Sparsity]
	 \addplot[line width=1.8pt, mark=*, mark size=3, color=red, mark=triangle, mark options={solid, draw=black}] coordinates {(0.94708,0.96010) (0.95632,0.96330) (0.97208,0.95940) (0.98257,0.96070) (0.98864,0.96140) (0.99215,0.95840)};
	 \addplot[line width=1.8pt, mark=*, mark size=3, color=blue, dashed, mark=o, mark options={solid, draw=black}] coordinates {(0.93961,0.95610) (0.94171,0.95610) (0.94603,0.95590) (0.95029,0.95420) (0.95441,0.95420) (0.95833,0.95630)};
	 \addplot[line width=1.8pt, mark=*, mark size=3, color=purple, mark=square, mark options={solid, draw=black}] coordinates {(0.94334,0.95310) (0.94929,0.95470) (0.96034,0.95610) (0.96977,0.95730) (0.97728,0.95660) (0.98283,0.95290)};
	 \addplot[line width=1.8pt, mark=*, mark size=3, color=green, dashed, mark=diamond, mark options={solid, draw=black}] coordinates {(0.94523,0.94400) (0.95282,0.94610) (0.96557,0.94700) (0.97344,0.94260) (0.97929,0.94920) (0.98388,0.94730)};
	 \addplot[line width=2pt, dotted, color=magenta] coordinates {(0.93961, 0.9842) (0.99215, 0.9842)};
	 \addplot[line width=2pt,  color=cyan] coordinates {(0.93961, 0.9586) (0.99215, 0.9586)};
	 \addplot[line width=2pt, color=brown] coordinates{(0.93961, 0.9498) (0.99215, 0.9498)};
	\end{axis}
	\end{tikzpicture}
	}~
	\subfloat[en]{
	\begin{tikzpicture}[scale=.48]
	\begin{axis}[xmax=1,xlabel=Sparsity, ymin=.955, ymax=.985]
	 \addplot[line width=1.8pt, mark=*, mark size=3, color=red, mark=triangle, mark options={solid, draw=black}] coordinates {(0.94902,0.97280) (0.95956,0.97200) (0.97494,0.97200) (0.98386,0.97250) (0.98910,0.97200) (0.99225,0.97230)};
	 \addplot[line width=1.8pt, mark=*, mark size=3, color=blue, dashed, mark=o, mark options={solid, draw=black}] coordinates {(0.93953,0.97060) (0.94152,0.97080) (0.94555,0.97010) (0.94950,0.97060) (0.95329,0.97040) (0.95689,0.97070)};
	 \addplot[line width=1.8pt, mark=*, mark size=3, color=purple, mark=square, mark options={solid, draw=black}] coordinates {(0.94354,0.96990) (0.94966,0.96910) (0.96099,0.96960) (0.97059,0.96940) (0.97820,0.96800) (0.98372,0.96770)};
	 \addplot[line width=1.8pt, mark=*, mark size=3, color=green, dashed, mark=diamond, mark options={solid, draw=black}] coordinates {(0.94374,0.96200) (0.95009,0.96100) (0.96152,0.96200) (0.96938,0.96170) (0.97539,0.96200) (0.98032,0.96140)};
	 \addplot[line width=2pt, dotted, color=magenta] coordinates {(0.93953, 0.9774) (0.99225, 0.9774)};
	 \addplot[line width=2pt,  color=cyan] coordinates {(0.93953, 0.9699) (0.99225, 0.9699)};
	 \addplot[line width=2pt, color=brown] coordinates{(0.93953, 0.9651) (0.99225, 0.9651)};
	\end{axis}
	\end{tikzpicture}
	}\\
	\subfloat[es]{
	\begin{tikzpicture}[scale=.48]
	\begin{axis}[xmax=1,xlabel=Sparsity,ylabel=Per token accuracy]
	 \addplot[line width=1.8pt, mark=*, mark size=3, color=red, mark=triangle, mark options={solid, draw=black}] coordinates {(0.94839,0.96030) (0.95871,0.96140) (0.97502,0.96100) (0.98490,0.96100) (0.99030,0.96140) (0.99333,0.96070)};
	 \addplot[line width=1.8pt, mark=*, mark size=3, color=blue, dashed, mark=o, mark options={solid, draw=black}] coordinates {(0.93985,0.95790) (0.94219,0.95750) (0.94698,0.95770) (0.95172,0.95380) (0.95619,0.95140) (0.96046,0.95800)};
	 \addplot[line width=1.8pt, mark=*, mark size=3, color=purple, mark=square, mark options={solid, draw=black}] coordinates {(0.94354,0.95310) (0.94966,0.95290) (0.96108,0.95360) (0.97100,0.95540) (0.97871,0.95450) (0.98424,0.95150)};
	 \addplot[line width=1.8pt, mark=*, mark size=3, color=green, dashed, mark=diamond, mark options={solid, draw=black}] coordinates {(0.94652,0.93570) (0.95524,0.93360) (0.96878,0.93710) (0.97667,0.93730) (0.98245,0.93270) (0.98680,0.93990)};
	 \addplot[line width=2pt, dotted, color=magenta] coordinates {(0.93985, 0.9643) (0.99333, 0.9643)};
	 \addplot[line width=2pt,  color=cyan] coordinates {(0.93985, 0.9231) (0.99333, 0.9231)};
	 \addplot[line width=2pt, color=brown] coordinates{(0.93985, 0.9568) (0.99333, 0.9568)};
	\end{axis}
	\end{tikzpicture}
	}~
	\subfloat[hu]{
	\begin{tikzpicture}[scale=.48]
	\begin{axis}[xmax=1,xlabel=Sparsity]
	 \addplot[line width=1.8pt, mark=*, mark size=3, color=red, mark=triangle, mark options={solid, draw=black}] coordinates {(0.95243,0.92910) (0.96582,0.92920) (0.98327,0.93040) (0.99077,0.92890) (0.99422,0.92200) (0.99607,0.92010)};
	 \addplot[line width=1.8pt, mark=*, mark size=3, color=blue, dashed, mark=o, mark options={solid, draw=black}] coordinates {(0.94140,0.92320) (0.94527,0.92170) (0.95291,0.91630) (0.96004,0.92060) (0.96633,0.92050) (0.97169,0.92100)};
	 \addplot[line width=1.8pt, mark=*, mark size=3, color=purple, mark=square, mark options={solid, draw=black}] coordinates {(0.94521,0.91730) (0.95255,0.91970) (0.96521,0.91870) (0.97490,0.91650) (0.98163,0.91640) (0.98606,0.91500)};
	 \addplot[line width=1.8pt, mark=*, mark size=3, color=green, dashed, mark=diamond, mark options={solid, draw=black}] coordinates {(0.95456,0.88550) (0.96711,0.88620) (0.98057,0.89070) (0.98726,0.89090) (0.99127,0.88880) (0.99385,0.88700)};
	 \addplot[line width=4pt, dotted, color=magenta] coordinates {(0.94140, 0.9536) (0.99607, 0.9536)};
	 \addplot[line width=2pt,  color=cyan] coordinates {(0.94140, 0.8633) (0.99607, 0.8633)};
	 \addplot[line width=2pt, color=brown] coordinates{(0.94140, 0.9054) (0.99607, 0.9054)};
	\end{axis}
	\end{tikzpicture}
	}~
	\subfloat[it]{
	\begin{tikzpicture}[scale=.48]
	\begin{axis}[xmax=1,xlabel=Sparsity]
	 \addplot[line width=1.8pt, mark=*, mark size=3, color=red, mark=triangle, mark options={solid, draw=black}] coordinates {(0.96037,0.94920) (0.97482,0.95210) (0.98772,0.95020) (0.99285,0.94720) (0.99533,0.94800) (0.99669,0.94230)};
	 \addplot[line width=1.8pt, mark=*, mark size=3, color=blue, dashed, mark=o, mark options={solid, draw=black}] coordinates {(0.93977,0.94510) (0.94202,0.94510) (0.94664,0.95020) (0.95120,0.94800) (0.95559,0.94900) (0.95977,0.94490)};
	 \addplot[line width=1.8pt, mark=*, mark size=3, color=purple, mark=square, mark options={solid, draw=black}] coordinates {(0.94340,0.94210) (0.94939,0.94110) (0.96059,0.94150) (0.97034,0.94210) (0.97801,0.94600) (0.98357,0.94390)};
	 \addplot[line width=1.8pt, mark=*, mark size=3, color=green, dashed, mark=diamond, mark options={solid, draw=black}] coordinates {(0.94621,0.92760) (0.95472,0.92880) (0.96792,0.91990) (0.97560,0.93010) (0.98121,0.92050) (0.98559,0.92450)};
	 \addplot[line width=2pt, dotted, color=magenta] coordinates {(0.93977, 0.9594) (0.99669, 0.9594)};
	 \addplot[line width=2pt,  color=cyan] coordinates {(0.94140, 0.8932) (0.99607, 0.8932)};
	 \addplot[line width=2pt, color=brown] coordinates{(0.93977, 0.9490) (0.99669, 0.9490)};
	\end{axis}
	\end{tikzpicture}
	}~
	\subfloat[nl]{
	\begin{tikzpicture}[scale=.48]
	\begin{axis}[xmax=1,xlabel=Sparsity]
	 \addplot[line width=1.8pt, mark=*, mark size=3, color=red, mark=triangle, mark options={solid, draw=black}] coordinates {(0.95996,0.93050) (0.97427,0.93430) (0.98823,0.92970) (0.99358,0.92880) (0.99594,0.92830) (0.99718,0.90210)};
	 \addplot[line width=1.8pt, mark=*, mark size=3, color=blue, dashed, mark=o, mark options={solid, draw=black}] coordinates {(0.94055,0.93390) (0.94357,0.92610) (0.94963,0.92560) (0.95539,0.92620) (0.96077,0.92500) (0.96563,0.92780)};
	 \addplot[line width=1.8pt, mark=*, mark size=3, color=purple, mark=square, mark options={solid, draw=black}] coordinates {(0.94426,0.92370) (0.95082,0.93120) (0.96263,0.92110) (0.97229,0.92560) (0.97958,0.92830) (0.98468,0.92970)};
	 \addplot[line width=1.8pt, mark=*, mark size=3, color=green, dashed, mark=diamond, mark options={solid, draw=black}] coordinates {(0.94881,0.90190) (0.95893,0.90160) (0.97307,0.90830) (0.98097,0.90950) (0.98625,0.90660) (0.98995,0.90740)};
	 \addplot[line width=2pt, dotted, color=magenta] coordinates {(0.94055, 0.9447) (0.99718, 0.9447)};
	 \addplot[line width=2pt,  color=cyan] coordinates {(0.94140, 0.8879) (0.99607, 0.8879)};
	 \addplot[line width=2pt, color=brown] coordinates{(0.94055, 0.9271) (0.99718, 0.9271)};
	\end{axis}
	\end{tikzpicture}
	}\\
	\subfloat[pt]{
	\begin{tikzpicture}[scale=.48]
	\begin{axis}[xmax=1,xlabel=Sparsity,ylabel=Per token accuracy]
	 \addplot[line width=1.8pt, mark=*, mark size=3, color=red, mark=triangle, mark options={solid, draw=black}] coordinates {(0.95941,0.95700) (0.97417,0.95960) (0.98831,0.95840) (0.99355,0.96060) (0.99588,0.95890) (0.99710,0.95360)};
	 \addplot[line width=1.8pt, mark=*, mark size=3, color=blue, dashed, mark=o, mark options={solid, draw=black}] coordinates {(0.94055,0.95450) (0.94358,0.95420) (0.94962,0.95480) (0.95537,0.95620) (0.96069,0.95450) (0.96550,0.95400)};
	 \addplot[line width=1.8pt, mark=*, mark size=3, color=purple, mark=square, mark options={solid, draw=black}] coordinates {(0.94422,0.95210) (0.95086,0.95280) (0.96288,0.95210) (0.97283,0.95110) (0.98021,0.95430) (0.98528,0.95300)};
	 \addplot[line width=1.8pt, mark=*, mark size=3, color=green, dashed, mark=diamond, mark options={solid, draw=black}] coordinates {(0.94932,0.94490) (0.95978,0.94650) (0.97358,0.94460) (0.98115,0.94200) (0.98630,0.94320) (0.98998,0.94510)};
	 \addplot[line width=2pt, dotted, color=magenta] coordinates {(0.94055, 0.9773) (0.99710, 0.9773)};
	 \addplot[line width=2pt,  color=cyan] coordinates {(0.94055, 0.9328) (0.99710, 0.9328)};
	 \addplot[line width=2pt, color=brown] coordinates{(0.94055, 0.9552) (0.99710, 0.9552)};
	\end{axis}
	\end{tikzpicture}
	}~
	\subfloat[sl]{
	\begin{tikzpicture}[scale=.48]
	\begin{axis}[xmax=1,xlabel=Sparsity]
	 \addplot[line width=1.8pt, mark=*, mark size=3, color=red, mark=triangle, mark options={solid, draw=black}] coordinates {(0.94517,0.94410) (0.95197,0.94100) (0.96432,0.94010) (0.97480,0.93870) (0.98299,0.93900) (0.98880,0.93740)};
	 \addplot[line width=1.8pt, mark=*, mark size=3, color=blue, dashed, mark=o, mark options={solid, draw=black}] coordinates {(0.94415,0.92710) (0.95068,0.92960) (0.96233,0.92990) (0.97124,0.93470) (0.97760,0.93210) (0.98221,0.93300)};
	 \addplot[line width=1.8pt, mark=*, mark size=3, color=purple, mark=square, mark options={solid, draw=black}] coordinates {(0.94984,0.92900) (0.96007,0.92630) (0.97456,0.92860) (0.98259,0.93430) (0.98725,0.93110) (0.99012,0.92820)};
	 \addplot[line width=1.8pt, mark=*, mark size=3, color=green, dashed, mark=diamond, mark options={solid, draw=black}] coordinates {(0.96553,0.91020) (0.97818,0.90310) (0.98834,0.90220) (0.99277,0.90380) (0.99529,0.90190) (0.99681,0.89500)};
	 \addplot[line width=2pt, dotted, color=magenta] coordinates {(0.94415, 0.9390) (0.99681, 0.9390)};
	 \addplot[line width=2pt,  color=cyan] coordinates {(0.94055, 0.8712) (0.99710, 0.8712)};
	 \addplot[line width=2pt, color=brown] coordinates{(0.94415, 0.9183) (0.99681, 0.9183)};
	\end{axis}
	\end{tikzpicture}
	}~
	\subfloat[sv]{
	\begin{tikzpicture}[scale=.48]
	\begin{axis}[xmax=1,xlabel=Sparsity]
	 \addplot[line width=1.8pt, mark=*, mark size=3, color=red, mark=triangle, mark options={solid, draw=black}] coordinates {(0.95593,0.94220) (0.96998,0.94360) (0.98557,0.93900) (0.99222,0.93860) (0.99519,0.93900) (0.99673,0.93050)};
	 \addplot[line width=1.8pt, mark=*, mark size=3, color=blue, dashed, mark=o, mark options={solid, draw=black}] coordinates {(0.94110,0.93790) (0.94467,0.93180) (0.95177,0.93480) (0.95842,0.93550) (0.96440,0.93410) (0.96970,0.93530)};
	 \addplot[line width=1.8pt, mark=*, mark size=3, color=purple, mark=square, mark options={solid, draw=black}] coordinates {(0.94521,0.93600) (0.95241,0.93600) (0.96509,0.93350) (0.97478,0.93140) (0.98159,0.93920) (0.98606,0.93560)};
	 \addplot[line width=1.8pt, mark=*, mark size=3, color=green, dashed, mark=diamond, mark options={solid, draw=black}] coordinates {(0.95131,0.92260) (0.96277,0.92190) (0.97659,0.91990) (0.98401,0.92010) (0.98883,0.92240) (0.99209,0.92680)};
	 \addplot[line width=2pt, dotted, color=magenta] coordinates {(0.94110, 0.9556) (0.99673, 0.9556)};
	 \addplot[line width=2pt,  color=cyan] coordinates {(0.94055, 0.9151) (0.99710, 0.9151)};
	 \addplot[line width=2pt, color=brown] coordinates{(0.94110, 0.9341) (0.99673, 0.9341)};
	\end{axis}
	\end{tikzpicture}
	}~
	\subfloat[tr]{
	\begin{tikzpicture}[scale=.48]
	\begin{axis}[xmax=1,xlabel=Sparsity]
	 \addplot[line width=1.8pt, mark=*, mark size=3, color=red, mark=triangle, mark options={solid, draw=black}] coordinates {(0.95328,0.85780) (0.96704,0.85930) (0.98431,0.86670) (0.99134,0.85850) (0.99449,0.85570) (0.99624,0.84630)};
	 \addplot[line width=1.8pt, mark=*, mark size=3, color=blue, dashed, mark=o, mark options={solid, draw=black}] coordinates {(0.94176,0.85760) (0.94601,0.85120) (0.95427,0.85170) (0.96175,0.85480) (0.96821,0.85770) (0.97350,0.85540)};
	 \addplot[line width=1.8pt, mark=*, mark size=3, color=purple, mark=square, mark options={solid, draw=black}] coordinates {(0.94586,0.85310) (0.95357,0.84990) (0.96664,0.85070) (0.97631,0.85210) (0.98287,0.84960) (0.98709,0.84720)};
	 \addplot[line width=1.8pt, mark=*, mark size=3, color=green, dashed, mark=diamond, mark options={solid, draw=black}] coordinates {(0.95611,0.82870) (0.96876,0.83360) (0.98169,0.83170) (0.98819,0.82830) (0.99202,0.82830) (0.99447,0.82960)};
	 \addplot[line width=2pt, dotted, color=magenta] coordinates {(0.94176, 0.8963) (0.99624, 0.8963)};
	 \addplot[line width=2pt,  color=cyan] coordinates {(0.94055, 0.8350) (0.99710, 0.8350)};
	 \addplot[line width=2pt, color=brown] coordinates{(0.94176, 0.8337) (0.99624, 0.8337)};
	\end{axis}
	\end{tikzpicture}
	}\\
	\ref{named}
	\caption{POS tagging results on the CoNLL 2006/07 treebanks evaluating against universal POS tags. Ticks are placed for $\lambda=0.05,0.1,0.2,0.3,0.4,0.5$. The x-axis shows the sparsity of the representations.}
	\label{fig:conllXresults}
\end{figure*}

\paragraph{Comparing word embeddings}

Our motivation for choosing \texttt{polyglot} word embeddings as input to sparse coding is that they are publicly available for a variety of languages. However, distributed word representations trained in any other reasonable manner can serve as input to our approach. In order to investigate if some of the popular word embedding techniques seem favorable for our algorithm, we conduct experiments using alternatively trained embeddings, i.e.~skip-gram (SG), continuous bag-of-words (CBOW) and Glove.

In order the utility of different word embeddings not to conflate with other factors, we train them on the same Wikipedia dumps used for training the \texttt{polyglot} word vectors. We choose further hyperparameters identically to \texttt{polyglot}, i.e.~we train 64 dimensional dense word representations using a symmetric context window of size 2 for both SG/CBOW\footnote{\url{https://code.google.com/archive/p/word2vec/}} and Glove\footnote{\url{http://nlp.stanford.edu/projects/glove/}}.

\begin{table*}[!ht]
\centering
\footnotesize
\subfloat[Results obtained using sparse word representations ($\lambda=0.1, m=1024$).]{
\begin{tabular}{@{}p{1.6cm}|ccccccccccccc}
 & bg & da & de & en & es & hu & it & nl & pt & sl & sv & tr & Avg. \\ \hline
\texttt{polyglot}\textsubscript{SC} & 96.04 & 95.71 & 96.33 & 97.20 & 96.14 & 92.92 & 95.21 & 93.43 & 95.96 & 94.10 & 94.36 & 85.93 & 94.44 \\
CBOW\textsubscript{SC}     & 95.10 & 95.35 & 95.61 & 97.08 & 95.75 & 92.17 & 94.51 & 92.61 & 95.42 & 92.96 & 93.18 & 85.12 & 93.74 \\
SG\textsubscript{SC}       & 94.67 & 95.49 & 95.47 & 96.91 & 95.29 & 91.97 & 94.11 & 93.12 & 95.28 & 92.63 & 93.60 & 84.99 & 93.63 \\
Glove\textsubscript{SC}    & 93.16 & 93.63 & 94.61 & 96.10 & 93.36 & 88.62 & 92.88 & 90.16 & 94.65 & 90.31 & 92.19 & 83.36 & 91.92
\end{tabular}
} \\

\subfloat[Results obtained using dense word representations.]{
\begin{tabular}{@{}p{1.6cm}|ccccccccccccc}
 & bg & da & de & en & es & hu & it & nl & pt & sl & sv & tr & Avg. \\ \hline
\texttt{polyglot} & 92.11 & 93.03 & 93.10 & 94.80 & 94.64 & 89.23 & 92.90 & 90.07 & 94.36 & 89.36 & 89.14 & 81.33 & 91.17 \\
CBOW     & 90.19 & 90.36 & 88.46 & 91.22 & 91.55 & 86.07 & 87.11 & 88.09 & 92.45 & 87.82 & 87.00 & 79.30 & 88.30 \\
SG       & 88.10 & 88.84 & 86.48 & 90.19 & 91.34 & 84.38 & 85.09 & 85.11 & 91.77 & 88.17 & 84.48 & 78.72 & 86.89 \\
Glove    & 83.10 & 81.95 & 83.07 & 86.64 & 84.65 & 77.34 & 79.98 & 78.54 & 86.62 & 80.91 & 78.77 & 76.77 & 81.53
\end{tabular}
\label{table:sparse_vs_denseB}
}
\caption{Performances of sparse and dense word representations for POS tagging over the 12 CoNLL-X datasets.}
\label{table:sparse_vs_dense}
\end{table*}

Figure~\ref{fig:conllXresults} includes POS tagging accuracies over the 12 treebanks from the CoNLL 2006/07 shared tasks evaluated against Google Universal POS tags. Instead of reporting results as a function of $\lambda$, we rather present accuracies as a function of the different sparsity levels induced by different $\lambda$ values. Figure~\ref{fig:conllXresults} demonstrates that POS tagging performance is quite insensitive to the choice of $\lambda$ unless it yields some extreme sparsity level ($>$99.5\%).

Figure~\ref{fig:conllXresults} also reveals that the usage of \texttt{polyglot}\textsubscript{SC} word representations tend to produce superior results over all alternative representations we experiment with. Furthermore, models using \texttt{polyglot}\textsubscript{SC} consistently outperform the FR\textsubscript{w} and Brown clustering-based baselines.

Models relying on SG\textsubscript{SC} and CBOW\textsubscript{SC} representations have an average tagging accuracy of 93.74 and 93.63, respectively, and they typically perform better than the baseline using Brown clustering with an average tagging performance of 93.27. Although utilizing Glove embeddings produce the lowest scores (91.92 on average), its scores still surpass those of the FR\textsubscript{w} baseline for all languages except for Turkish.

The average tagging performance over the 12 languages when relying on features based on \texttt{polyglot}\textsubscript{SC} is only 1.3 points below that of $FR_{w+c}$ (i.e.~94.4 versus 95.7). Recall that $FR_{w+c}$ uses a feature-rich representation, whereas our proposed model uses only $O(m)$ features, i.e.~it is tied to the number of the basis vectors employed for sparse coding. Furthermore, neither does our model employs word identity features nor it relies on character-level features of words.

\paragraph{Analyzing the effects of window size}

Hyperparameters for training word representations can largely impact their quality as also concluded by \newcite{DBLP:journals/tacl/LevyGD15}. We thus investigate if providing a larger context window size during the training of CBOW, SG and Glove embeddings can improve their utility of being employed in our model.

\begin{figure}
\includegraphics[width=\columnwidth]{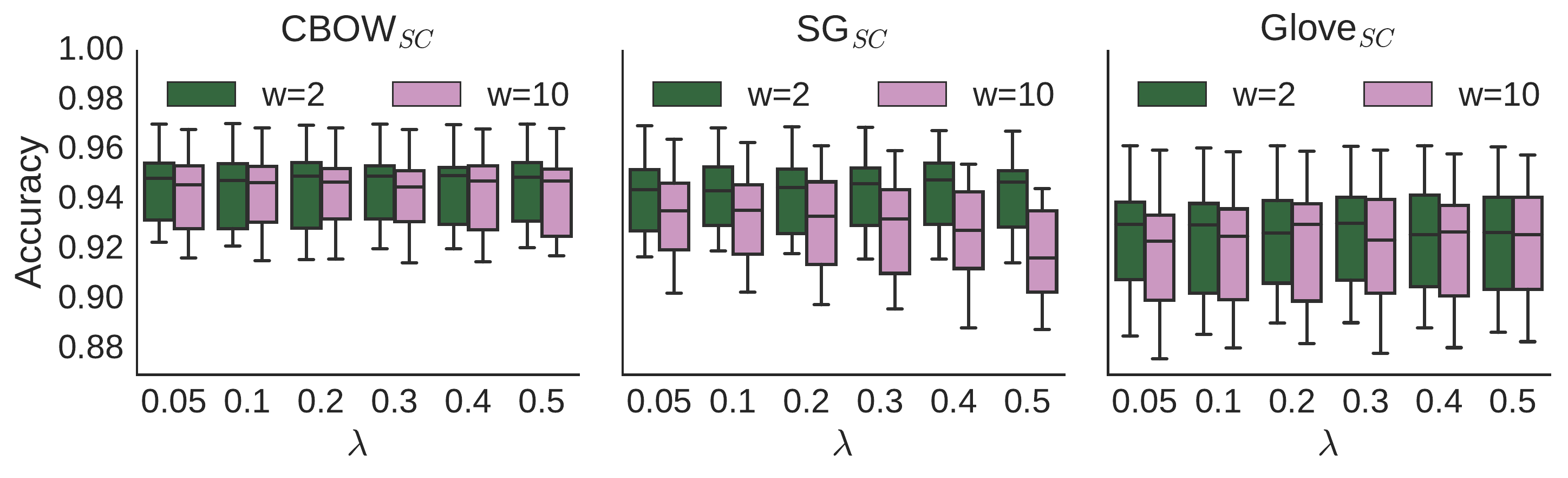}
\caption{Overview of POS tagging accuracies over the 12 CoNLL-X datasets when relying on sparse coded versions of alternative word embeddings trained with context window size of 2 and 10.}
\label{fig:window_size_box}
\end{figure}

According to Figure~\ref{fig:window_size_box} applying context window sizes of 2 for training the word embeddings tend to produce better overall POS tagging accuracies than applying a larger window size of 10. Differences are the most pronounced in case of skip-gram representation, confirming the findings of \newcite{lin-EtAl:2015:NAACL-HLT}, i.e.~embedding models that model short-range context are more effective for POS tagging.

\paragraph{Comparing dense and sparse representations}

In accordance to Figure~\ref{fig:conllXresults}, we set the value of $\lambda$ to $0.1$ for our upcoming experiments unless stated otherwise. Table~\ref{table:sparse_vs_dense} demonstrates that performances obtained by models using dense word representations as features are consistently inferior to those models which enjoy the benefit of features derived from sparse word representations.

\begin{table*}[!ht]
\centering
\footnotesize
\subfloat[Results obtained with different models when all the training corpora was used.]{
\begin{tabular}{@{}p{1.6cm}|ccccccccccccc}
model  &   bg  &   da  &   de  &   en  &   es  &   hu  &   it  &   nl  &   pt  &   sl  &   sv  &   tr  & Avg. \\ \hline
\texttt{polyglot}\textsubscript{SC} & 96.04 & 95.71 & 96.33 & 97.20 & 96.14 & 92.92 & 95.21 & 93.43 & 95.96 & \textbf{94.10} & 94.36 & 85.93 & 94.44 \\
FR\textsubscript{w} & 92.55 & 91.68 & 95.86 & 96.99 & 92.31 & 86.33 & 89.32 & 88.79 & 93.28 & 87.12 & 91.51 & 83.50 & 90.77\\
FR\textsubscript{w+c} & \textbf{97.20} & \textbf{96.67} & \textbf{98.42} & \textbf{97.74} & \textbf{96.43} & \textbf{95.36} & \textbf{95.94} & \textbf{94.47} & \textbf{97.73} & 93.90 & \textbf{95.56} & \textbf{89.63} & \textbf{95.75} \\ \hline
\#train sents. &12823 &5190 & 39216 & 39832 & 3306  & 6035  & 3110  & 13349 & 9071  & 1534  & 11042 & 4997  & 12458\\
\end{tabular}
} \\

\subfloat[Results obtained with different models when the first 1,500 sentences of the training corpora were used.]{
\begin{tabular}{@{}p{1.6cm}|ccccccccccccc}
model  &  bg  &   da  &   de  &   en  &   es  &   hu  &   it  &   nl  &   pt  &   sl  &   sv  &   tr  & Avg. \\ \hline
\texttt{polyglot}\textsubscript{SC} & 88.20 & \textbf{94.04} & 93.47 & \textbf{95.76} & \textbf{95.63} & 91.15 & \textbf{94.19} & \textbf{87.28} & 94.60 & \textbf{94.12} & \textbf{91.14} & 83.23 & \textbf{91.90} \\
FR\textsubscript{w} & 79.63 & 87.75 & 85.58 & 90.93 & 89.87 & 80.01 & 86.60 & 74.40 & 89.13 & 86.93 & 80.16 & 77.59 & 85.05\\
FR\textsubscript{w+c} & \textbf{88.71} & 93.52 & \textbf{95.77} & 94.59 & 95.42 & \textbf{92.74} & 93.66 & 84.94 & \textbf{95.13} & 93.82 & 88.56 & \textbf{84.92} & 91.82 \\ \hline
train~sents.~\% & 11.70 & 28.90 & 3.82 & 3.77 & 45.37 & 24.86 & 48.23 & 11.24 & 16.54 & 97.78 & 13.58 & 30.02 & 12.04\\
\end{tabular}
} \\

\subfloat[Results obtained with different models when the first 150 sentences of the training corpora were used.]{
\begin{tabular}{@{}p{1.6cm}|ccccccccccccc}
model  &  bg  &   da  &   de  &   en  &   es  &   hu  &   it  &   nl  &   pt  &   sl  &   sv  &   tr  & Avg. \\ \hline
\texttt{polyglot}\textsubscript{SC} & \textbf{76.46} & \textbf{89.51} & 88.29 & \textbf{90.46} & \textbf{91.32} & \textbf{86.51} & \textbf{89.13} & \textbf{75.24} & \textbf{90.74} & \textbf{86.67} & \textbf{82.50} & \textbf{71.17} & \textbf{84.83} \\
FR\textsubscript{w}   & 62.44 & 74.88 & 72.46 & 78.10 & 77.80 & 67.20 & 75.45 & 56.67 & 79.38 & 72.46 & 65.13 & 61.38 & 70.28 \\
FR\textsubscript{w+c} & 74.87 & 83.34 & \textbf{89.64} & 85.75 & 85.88 & 83.54 & 84.99 & 69.28 & 87.52 & 83.88 & 76.71 & 67.40 & 81.07 \\ \hline
train~sents.~\% &1.17 & 2.89 & 0.38 & 0.38 & 4.54  & 2.49  & 4.82 & 1.12 & 1.65 & 9.78 & 1.36 & 3.00  & 1.20 \\
\end{tabular}
} \\
\caption{Comparison of models based on different amount of training data. Bold numbers indicate the best results for a given training regime (i.e.~either training on 150/1,500/all training sentences). \texttt{polyglot}\textsubscript{SC} uses $m=1024, \lambda=0.1$.}
\label{table:training_size}
\end{table*}

In Table~\ref{table:sparse_vs_denseB}, we can see that \texttt{polyglot} embeddings perform the best for dense representations as well. When using dense features, the CBOW representation-based model tends to produce results better by a 1.4 points margin on average compared to SG embeddings. This performance gap between the two \texttt{word2vec} variants vanishes, however, when dense word representations are replaced by their sparse counterparts. Table~\ref{table:sparse_vs_dense} also reveals that sparse word representations improve average POS tagging accuracies by 3.3, 5.4, 6.7 and 10.4 points for \texttt{polylgot}, CBOW, SG and Glove word representations, respectively.

\paragraph{Comparing the effects of training corpus size}

\begin{figure}
\centering
\includegraphics[width=.7\columnwidth]{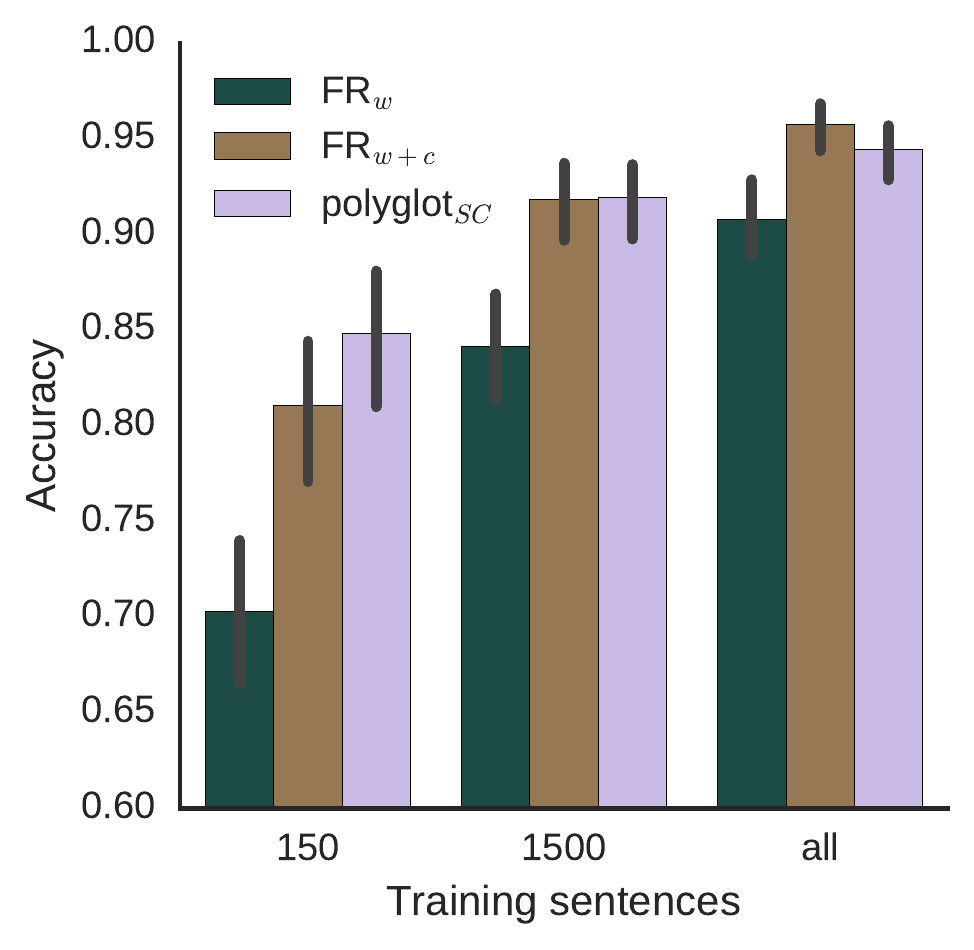}
\caption{Average tagging accuracies of different models over the 12 CoNLL-X languages using varying amount of training sentences.}
\label{fig:CONLLX_training_size_bar}
\end{figure}

We also investigate the generalization characteristics of the proposed representation by training models which have access to substantially different amounts of training data per language. We distinguish three scenarios, i.e.~when using only the first \textbf{150}, the first \textbf{1,500} and \textbf{all} the available training sentences from each corpora. Figure~\ref{fig:CONLLX_training_size_bar} illustrates the average POS tagging accuracy over the 12 CoNLL-X datasets for different amounts of training data and models.

Table~\ref{table:training_size} includes performances in more details revealing that the average performance of \texttt{polyglot}\textsubscript{SC} is 14.55 and 3.76 points better compared to the FR\textsubscript{w} and FR\textsubscript{w+c} baselines when using only 1.2\% of all the available training data, i.e.~150 sentences per each language. By discarding 98.8\% of the training data \texttt{polyglot}\textsubscript{SC} obtains 89.8\% of its average performance compared to the scenario when it has access to all the training sentences. However, under the same scenario the FR\textsubscript{w+c} and FR\textsubscript{w} models only manage to preserve 85\% and 77\% of their original performance, respectively.

Our model performs on par with FR\textsubscript{w+c} and has a 6.85 points advantage over FR\textsubscript{w} with a training corpus of 1,500 sentences. FR\textsubscript{w+c} has an average of 1.3 points advantage over \texttt{polyglot}\textsubscript{SC} when we provide access to all training data during training, nevertheless FR\textsubscript{w} still underperforms \texttt{polyglot}\textsubscript{SC} in that setting by 3.67 points.

\paragraph{Comparing sparse coding techniques}

Next, we compare different sparse coding approaches on the pre-trained \texttt{polyglot} word representations. The recent work of \newcite{faruqui-EtAl:2015:ACL-IJCNLP} formulated alternative approaches to determine sparse word representations. One of the objective functions \newcite{faruqui-EtAl:2015:ACL-IJCNLP} apply is
\begin{equation}
\min\limits_{D, \alpha} \frac{1}{2n} \sum_{i=1}^{n} \lVert \mathbf{x}_i-D \boldsymbol{\alpha}_i \rVert_2^2 + \lambda \lVert \boldsymbol{\alpha}_i \rVert_1 + \tau \lVert D \rVert_2^2.
\label{CMU_objective}
\end{equation}
The main difference in Eq.~\ref{SPAMS_objective} and \ref{CMU_objective} is that the latter does not explicitly constrain $D$ to be a member of the convex set of matrices comprising of column vectors having a pre-defined upper bound on their norm. In order to implicitly control for the norms of the basis vectors \newcite{faruqui-EtAl:2015:ACL-IJCNLP} apply an additional regularization term affected by an extra parameter $\tau$ in their objective function.

\newcite{faruqui-EtAl:2015:ACL-IJCNLP} also formulated a constrained objective function of the form
\begin{equation}
\small
\min\limits_{\substack{D \in \mathbb{R}^{k \times m}_{\geq 0} \\  \alpha \in \mathbb{R}^{k \times \lvert V \rvert}_{\geq 0}}} \frac{1}{2n} \sum_{i=1}^{n} \lVert \mathbf{x}_i-D \boldsymbol{\alpha}_i \rVert_2^2 + \lambda \lVert \boldsymbol{\alpha}_i \rVert_1 + \tau \lVert D \rVert_2^2,
\label{CMUNN_objective}
\end{equation}
for which a non-negativity constraint on the elements of $\alpha$ (but no constraint on $D$) is imposed. When using the objective functions introduced by \newcite{faruqui-EtAl:2015:ACL-IJCNLP}, we use the default $\tau=10^{-5}$ value. Notationally, we distinguish the sparse coding approaches based on the equation they use as their objective function, i.e.~SC-\textit{i}, $i\in\{1,3,4\}$.

\begin{figure}
\centering
\includegraphics[width=\columnwidth]{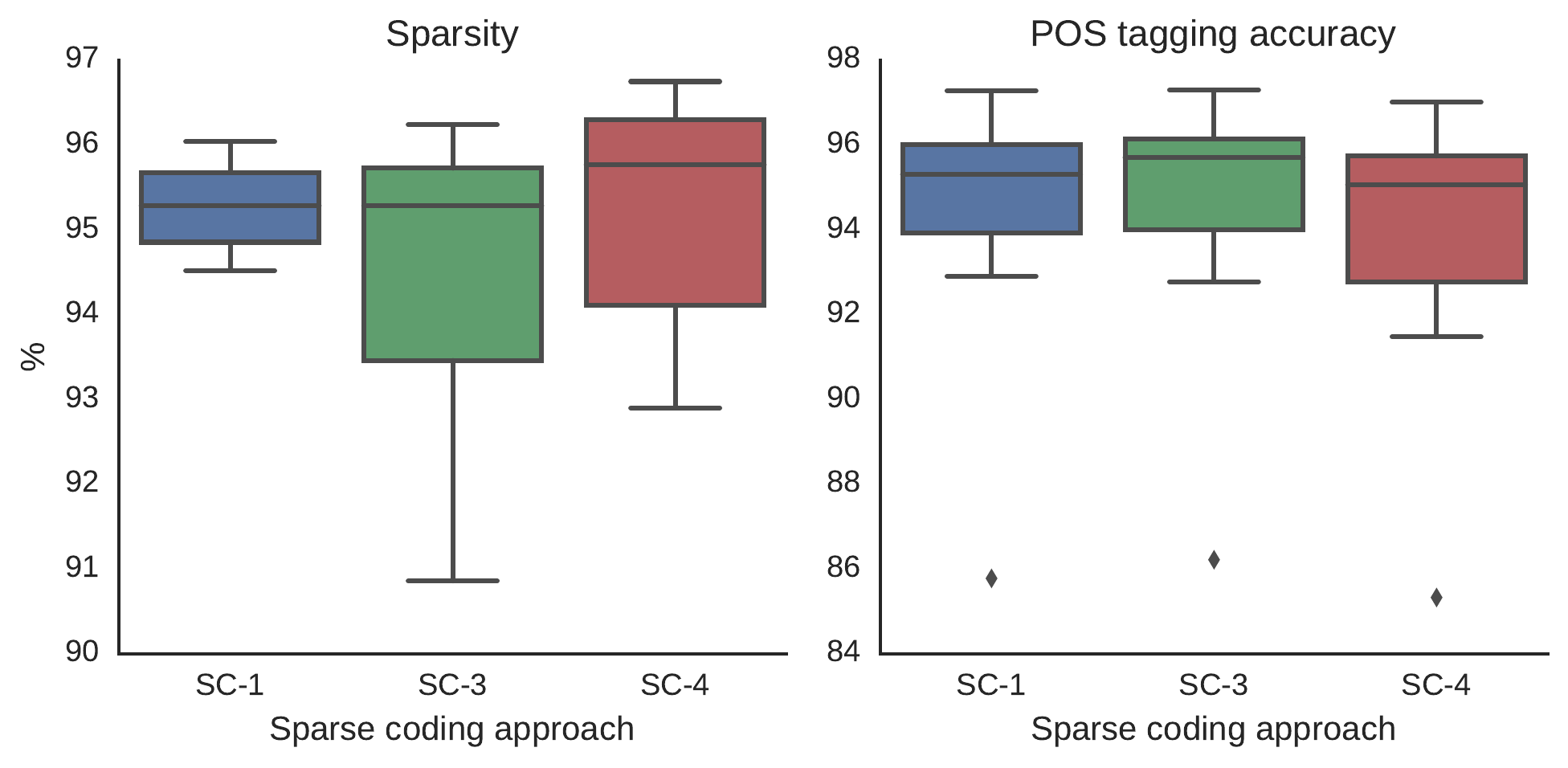}
\caption{Comparison of the POS tagging accuracies of different sparse coding techniques with comparable average sparseness levels over 12 CoNLL-X datasets.}
\label{SPAMS_VS_CMU}
\end{figure}

We applied $\lambda=0.05$ for SC-1 and $\lambda=0.5$ for SC-3 and SC-4 in order to obtain word representations of comparable average sparsity levels across the 12 languages, i.e.~95.3\%, 94.5\% and 95.2\%, respectively (cf.~the left of Figure~\ref{SPAMS_VS_CMU}).
The right of Figure~\ref{SPAMS_VS_CMU} further illustrates the spread of POS tagging accuracies over the 12 CoNLL-X treebanks when using models that rely on different sparse coding strategies with comparable sparsity levels.

Although \newcite{murphy-talukdar-mitchell:2012:PAPERS} mentions non-negativity as a desired property of word representations for cognitive plausibility, Figure~\ref{SPAMS_VS_CMU} reveals that our sequence labeling model cannot benefit from it as the average POS tagging accuracy for SC-4 is 0.7 points below that of SC-3 approach. The average performances when applying SC-1 and SC-3 are nearly identical with a 0.18 point difference between the two.

It is instructive to analyze the patterns different sparse coding approaches exhibit. Even though the objective functions used by the different approaches are similar, decompositions obtained by them convey rather different sparsity structures.

\begin{figure}
\centering
\subfloat[$\ell_2$ norms]{\includegraphics[width=\columnwidth]{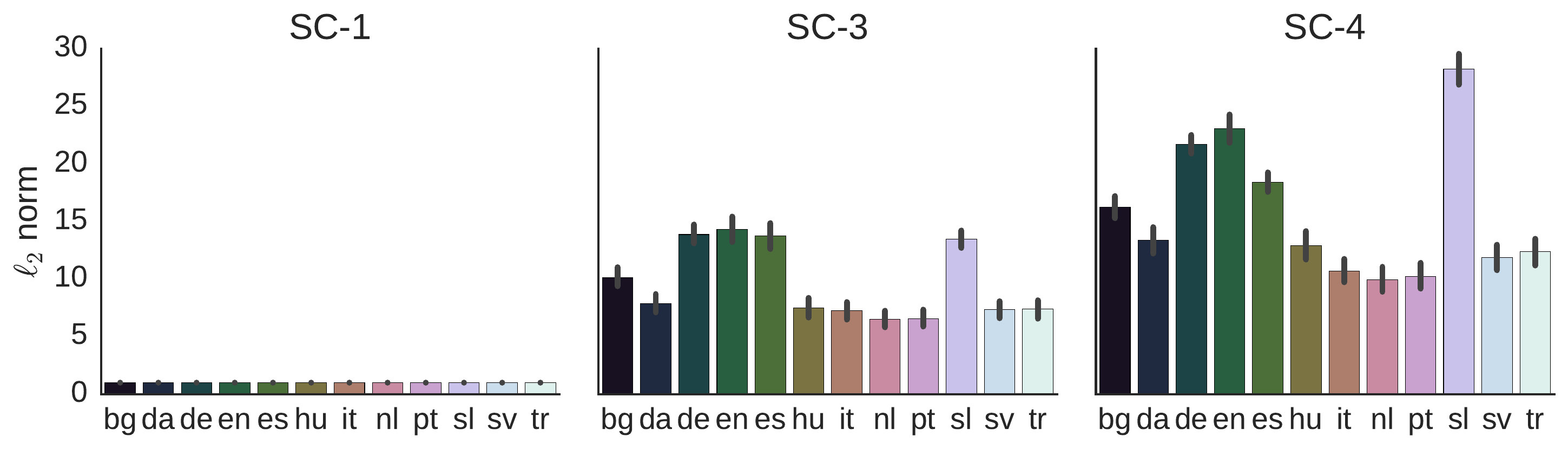} \label{fig:norms_vs_freqs2_A} } \\
\subfloat[Relative frequencies]{\includegraphics[width=\columnwidth]{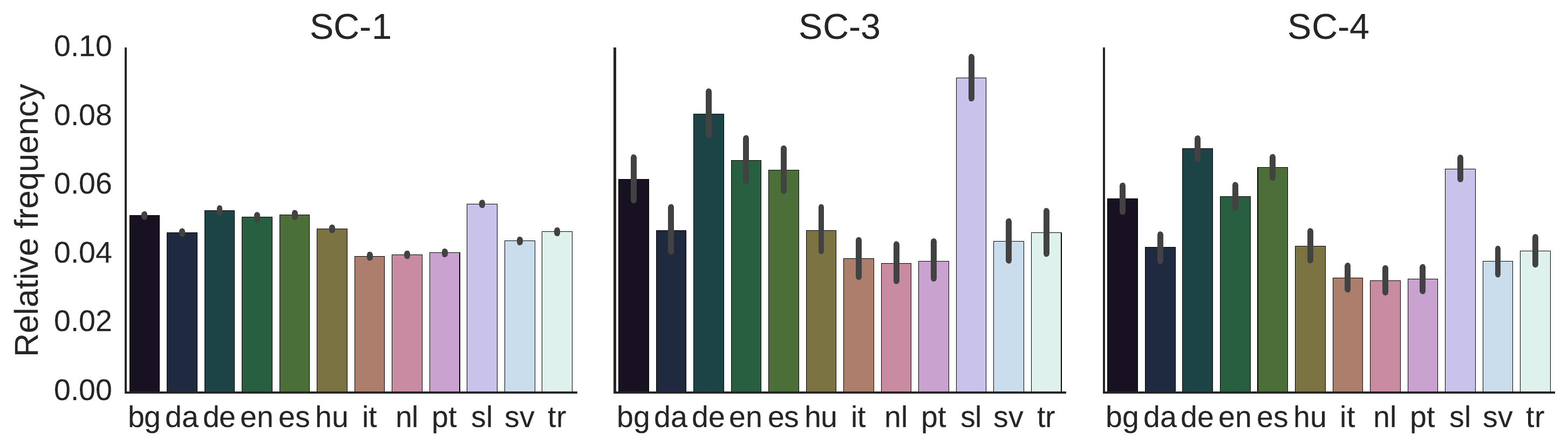} \label{fig:norms_vs_freqs2_B}} \\
\caption{Characteristics of the different sparse coding techniques over the 12 CoNLL-X languages.}
\label{fig:norms_vs_freqs2}
\end{figure}

Figure~\ref{fig:norms_vs_freqs2_A} illustrates that there exist substantial variation in the length of the basis vectors obtained by SC-3 and SC-4 both within and across languages. On the contrary, SC-1 produces practically no variation in the length of the basis vectors comprising $D$ due to the constraint present in the objective function it employs. Figure~\ref{fig:norms_vs_freqs2_B} shows similar differences about the relative frequency of basis vectors taking part in the reconstruction of word embeddings.

\begin{figure}
\centering
\includegraphics[width=.9\columnwidth]{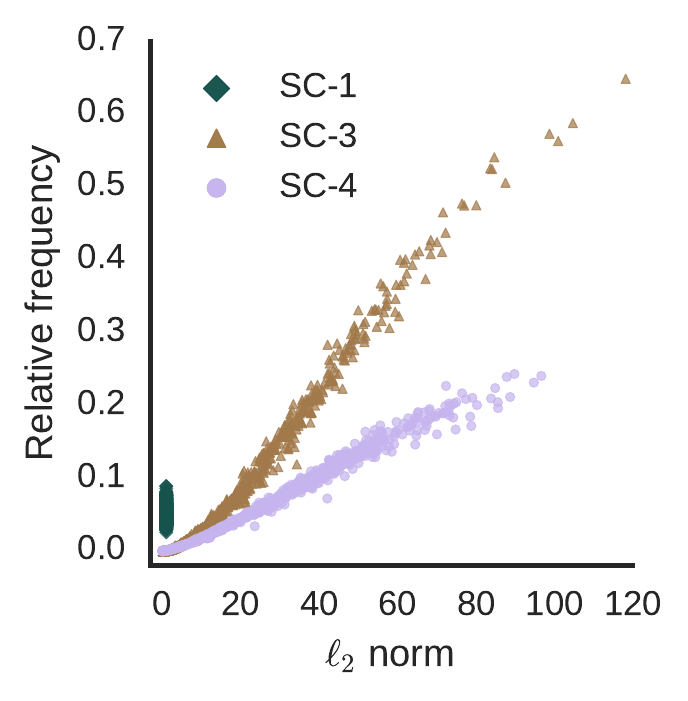}
\caption{Relative frequency of basis vectors receiving nonzero coefficients in $\alpha$ as a function of their $\ell_2$ norm.}
\label{fig:norms_vs_freqs_en}
\end{figure}

A further characteristic depicted in Figure~\ref{fig:norms_vs_freqs_en} is that strong correlation can be observed between the $\ell_2$ norm of basis vectors and the number of times they are assigned a non-zero coefficient in $\alpha$ for SC-3 and SC-4 but not for SC-1.

It can be further noted from Figure~\ref{fig:norms_vs_freqs_en} that the norm of the basis vectors determined by SC-3 and SC-4 are often orders of magnitude larger than those determined by SC-1. This effect, however, can be naturally mitigated by increasing $\tau$.

Overall, the different approaches convey comparable POS tagging accuracies but different decompositions due to the differences in the objective functions they employ. Upcoming experiments are conducted using the objective function in Eq.~\ref{SPAMS_objective}.

\subsubsection{Experiments using UD treebanks}

\begin{table*}[!ht]
\small
\centering
\begin{tabular}{l|cc|ccc|ccc||c}
 & \multicolumn{5}{c|}{Baseline using} & Word & Sparse &  & \\
& \multicolumn{2}{c|}{Words and characters} & \multicolumn{3}{c|}{Words only} & Identity & coding &  & Token\\
Treebank & bi-LSTM\textsubscript{w+c} & FR\textsubscript{w+c} & bi-LSTM\textsubscript{w} & FR\textsubscript{w} & Brown & (WI)  & (SC) &  WI+SC  &  coverage \\ \hline
\rowcolor{lightgray}bg & 98.25 & 96.88 & 95.12 & 90.40 & 93.36 & 90.75 & \textbf{95.33} & \textbf{95.63} & 92.64 \\
\rowcolor{lightgray}cs & 97.93 & 98.03 & 93.77 & 93.09 & 91.98 & 93.40 & \textbf{95.13} & \textbf{95.83} & 92.42 \\
\rowcolor{lightgray}da & 95.94 & 94.70 & 91.96 & 87.41 & 92.45 & 87.51 & \textbf{93.32} & \textbf{93.29} & 93.96 \\
\rowcolor{lightgray}de & 93.11 & 91.73 & 90.33 & 85.73 & 88.52 & 85.90 &      89.11     & \textbf{90.73} & 92.75 \\
el &  ---  & 96.77 &  ---  & 90.91 & 95.96 & 91.53 & \textbf{96.91} & \textbf{97.12} & 95.80 \\
\rowcolor{lightgray}en & 94.61 & 93.52 & 92.10 & 89.28 & 91.40 & 89.36 & \textbf{93.03} & \textbf{93.47} & 97.61 \\
\rowcolor{lightgray}es & 95.34 & 94.37 & 93.60 & 90.93 & 93.83 & 91.31 & \textbf{94.43} & \textbf{94.69} & 97.08 \\
et &  ---  & 84.83 &  ---  & 75.42 & 84.52 & 76.78 & \textbf{85.56} & \textbf{86.30} & 80.40 \\
eu & 94.91 & 93.03 & 88.00 & 83.36 &  ---  & 84.83 & \textbf{90.19} & \textbf{90.63} & 90.98 \\
\rowcolor{lightgray}fa & 96.89 & 96.13 & 95.31 & 93.98 & 95.04 & 94.45 & \textbf{95.91} & \textbf{96.11} & 97.80 \\
\rowcolor{lightgray}fi & 95.18 & 92.93 & 87.95 & 82.31 & 85.98 & 83.17 & \textbf{88.80} & \textbf{89.19} & 84.37 \\
fi\_ftb &---& 91.84&  ---  & 86.91 & 82.86 & 81.57 & \textbf{86.91} & \textbf{87.88} & 83.92 \\
\rowcolor{lightgray}fr & 96.04 & 95.30 & 94.44 & 92.80 & 92.42 & 92.88 &      93.52     & \textbf{94.96} & 92.06 \\
ga &  ---  & 89.64 &  ---  & 84.32 &  ---  & \textbf{85.21} & \textbf{88.22} & \textbf{88.82} & 88.80 \\
grc & ---  & 93.57 &  ---  & 84.35 & 57.13 & \textbf{84.44} & 70.27 & \textbf{85.04} & 43.58 \\
grc\_proiel&---&96.39& --- & 90.73 & 49.41 & \textbf{91.01} & 67.17 & \textbf{91.38} & 45.74 \\
\rowcolor{lightgray}he & 95.92 & 93.91 & 93.37 & 90.17 & 93.79 & 90.33 & \textbf{94.38} & \textbf{95.28} & 92.03 \\
\rowcolor{lightgray}hi & 96.64 & 95.96 & 95.99 & 94.32 & 94.61 & 94.25 &      95.37     & \textbf{96.09} & 96.40 \\
\rowcolor{lightgray}hr & 95.59 & 94.18 & 89.24 & 82.91 & 92.22 & 83.52 & \textbf{92.85} & \textbf{93.53} & 92.45 \\
hu &  ---  & 92.88 &  ---  & 73.69 & 91.08 & 75.63 &      89.47     &      89.47     & 90.07 \\
\rowcolor{lightgray}id & 92.79 & 93.32 & 90.48 & 87.29 & 91.39 & 88.03 & \textbf{91.71} & \textbf{92.02} & 97.09 \\
\rowcolor{lightgray}it & 97.64 & 96.92 & 96.57 & 93.62 & 94.92 & 93.43 &      95.70     &      96.28     & 94.99 \\
la &  ---  & 92.03 &  ---  & 77.75 &  ---  & \textbf{79.99} & \textbf{85.49} & \textbf{86.34} & 83.03 \\
la\_itt&---& 98.78 &  ---  & 97.69 &  ---  & \textbf{97.74} & 95.43 & \textbf{97.77} & 92.23 \\
la\_proiel&---&95.89& ---  & 90.53 &  ---  & \textbf{90.84} & 90.14 & \textbf{92.42} & 85.21 \\
\rowcolor{lightgray}nl & 92.07 & 88.79 & 84.96 & 81.11 & 84.28 & 81.27 &      84.32     & \textbf{85.10} & 92.28 \\
\rowcolor{lightgray}no & 97.77 & 96.53 & 94.39 & 91.58 & 94.29 & 91.87 & \textbf{95.42} & \textbf{95.67} & 94.53 \\
\rowcolor{lightgray}pl & 96.62 & 95.27 & 89.73 & 84.41 & 91.13 & 84.57 & \textbf{93.57} & \textbf{93.95} & 94.19 \\
\rowcolor{lightgray}pt & 97.48 & 96.59 & 94.24 & 90.69 & 93.74 & 91.11 &      94.00     & \textbf{95.50} & 92.53 \\
ro &  ---  & 86.46 &  ---  & 76.32 & 89.93 & 75.96 &      88.99     &      88.27     & 93.06 \\
\rowcolor{lightgray}sl & 97.78 & 95.28 & 91.09 & 84.43 & 90.24 & 84.92 & \textbf{92.65} & \textbf{92.70} & 92.14 \\
\rowcolor{lightgray}sv & 96.30 & 94.94 & 93.32 & 88.84 & 93.50 & 88.94 & \textbf{94.46} & \textbf{94.62} & 92.50 \\
ta &  ---  & 85.37 &  ---  & 68.02 &  ---  & \textbf{70.69} & \textbf{81.25} & \textbf{81.80} & 85.35 \\ \hline
Avg. & 95.99 & 94.76 & 92.40 & 88.77 & 91.95 & 89.05 & 93.15 & 93.73 & 93.59

\end{tabular}
\caption{Per token POS tagging accuracies for 33 UD treebanks. For sparse coding SPAMS is used on \texttt{polyglot} vectors with $\lambda=0.1$ and $m=1024$. Results in bold are better than any of bi-LSTM\textsubscript{w}, FR\textsubscript{w} and Brown models (i.e.~the baselines using features based on words only). Average is calculated over the 20 highlighted treebanks for which there are results in every column. The bi-LSTM results are from~\protect\newcite{plank-sogaard-goldberg:2016:P16-2}.}
\label{conllU}
\end{table*}

For POS tagging we also experiment with UD~v1.2 \cite{11234/1-1548} treebanks.
We used the default train-test splits of the treebanks not utilizing the development sets for fine tuning performance on any of the languages during our experiments. We omitted the Japanese treebank as words in it are stripped off due to licensing issues. Also there is no \texttt{polyglot} vector released for Old Church Slavonic and Gothic. Even though \texttt{polyglot} word representations are released for Arabic, it was of no practical use as it contained unvocalized surface forms of tokens in contrast to the vocalized forms in UD~v.1.2. For this reason, we discarded the Arabic treebank as less than~30\% of its tokens could be associated with a representation. By omitting these 4 languages from our experiments we are finally left with 33 treebanks for 29 languages. We note that for Ancient Greek treebanks (grc*) we use word embeddings trained on Modern Greek.

We should add that there are 4 languages (related to 6 treebanks) for which \texttt{polyglot} word vectors are accessible, however, the Wikipedia dumps used for training them are not distributed. For this reason, Brown clustering-based baselines are missing for the affected treebanks.

We report our results on UD~v1.2 in~Table~\ref{conllU}. Recall that the default behavior of our sparse coding-based models (SC in Table~\ref{conllU}) is that they do not handle word identity as an explicit feature. We now investigate how much contribution word identity features convey on their own and also when used in conjunction with sparse coding-derived features. For this end we introduce a simple linear chain CRF model generating features solely on the identity of the current word and the ones surrounding it (WI in Table~\ref{conllU}). Likewise, we define a model that relies on WI and SC features simultaneously (WI+SC). Table~\ref{conllU} reveals that SC outperforms WI by a large margin and that combining the two feature sets together yields some further improvements over SC scores.

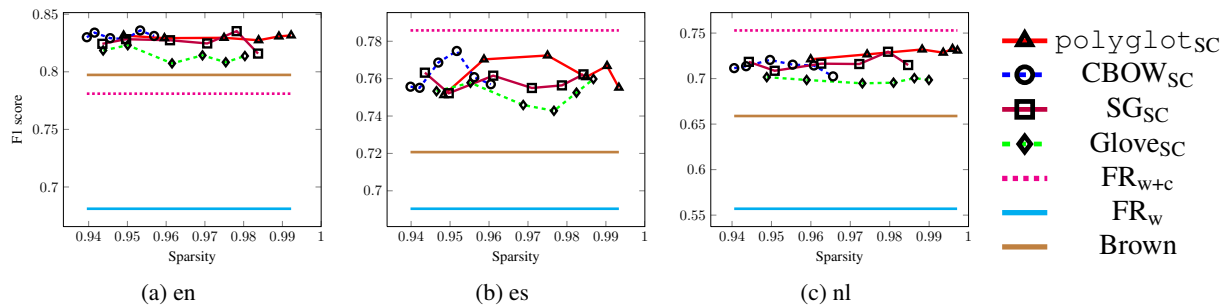
\begin{figure*}
\subfloat[en]{
\begin{tikzpicture}[scale=.5]
    \begin{axis}[xmax=1,xlabel=Sparsity,ylabel=F1 score,legend columns=1,legend to name=namedNER,legend entries={\texttt{polyglot}\textsubscript{SC},CBOW\textsubscript{SC},SG\textsubscript{SC},Glove\textsubscript{SC},FR\textsubscript{w+c},FR\textsubscript{w},Brown}, legend style={draw=none}]
 \addplot[line width=1.8pt, mark=*, mark size=3, color=red, mark=triangle, mark options={solid, draw=black}] coordinates {(0.94902,0.8315) (0.95956,0.8292) (0.97494,0.8295) (0.98386,0.8274) (0.98910,0.8307) (0.99225,0.8316)};
\addplot[line width=1.8pt, mark=*, mark size=3, color=blue, dashed, mark=o, mark options={solid, draw=black}] coordinates {(0.93953,0.8300) (0.94152,0.8340) (0.94555,0.8292) (0.94950,0.8284) (0.95329,0.8359) (0.95689,0.8311)};
\addplot[line width=1.8pt, mark=*, mark size=3, color=purple, mark=square, mark options={solid, draw=black}] coordinates {(0.94354,0.8243) (0.94966,0.8283) (0.96099,0.8274) (0.97059,0.8246) (0.97820,0.8353) (0.98372,0.8158)};
 \addplot[line width=1.8pt, mark=*, mark size=3, color=green, dashed, mark=diamond, mark options={solid, draw=black}] coordinates {(0.94374,0.8185) (0.95009,0.8231) (0.96152,0.8073) (0.96938,0.8143) (0.97539,0.8083) (0.98032,0.8137)};
 \addplot[line width=2pt, dotted, color=magenta] coordinates {(0.93953, 0.7810) (0.99225, 0.7810)};
 \addplot[line width=2pt,  color=cyan] coordinates {(0.93953, 0.6811) (0.99225, 0.6811)};
 \addplot[line width=2pt,  color=brown] coordinates {(0.93953, 0.7973) (0.99225, 0.7973)};
\end{axis}
\end{tikzpicture}
} \hfil
\subfloat[es]{
\begin{tikzpicture}[scale=.5]
\begin{axis}[xmax=1,xlabel=Sparsity] \addplot[line width=1.8pt, mark=*, mark size=3, color=red, mark=triangle, mark options={solid, draw=black}] coordinates {(0.94839,0.7512) (0.95871,0.7703) (0.97502,0.7724) (0.98490,0.7608) (0.99030,0.7668) (0.99333,0.7552)};
 \addplot[line width=1.8pt, mark=*, mark size=3, color=blue, dashed, mark=o, mark options={solid, draw=black}] coordinates {(0.93985,0.7557) (0.94219,0.7551) (0.94698,0.7686) (0.95172,0.7748) (0.95619,0.7608) (0.96046,0.7570)};
 \addplot[line width=1.8pt, mark=*, mark size=3, color=purple, mark=square, mark options={solid, draw=black}] coordinates {(0.94354,0.7632) (0.94966,0.7522) (0.96108,0.7616) (0.97100,0.7550) (0.97871,0.7564) (0.98424,0.7623)};
 \addplot[line width=1.8pt, mark=*, mark size=3, color=green, dashed, mark=diamond, mark options={solid, draw=black}] coordinates {(0.94652,0.7533) (0.95524,0.7578) (0.96878,0.7459) (0.97667,0.7428) (0.98245,0.7525) (0.98680,0.7598)};
 \addplot[line width=2pt, dotted, color=magenta] coordinates {(0.93985, 0.7858) (0.99333, 0.7858)};
 \addplot[line width=2pt,  color=cyan] coordinates {(0.93985, 0.6904) (0.99333, 0.6904)};
 \addplot[line width=2pt,  color=brown] coordinates {(0.93985, 0.7207) (0.99333, 0.7207)};
\end{axis}
\end{tikzpicture}
} \hfil
\subfloat[nl]{
\begin{tikzpicture}[scale=.5]
\begin{axis}[xmax=1,xlabel=Sparsity] \addplot[line width=1.8pt, mark=*, mark size=3, color=red, mark=triangle, mark options={solid, draw=black}] coordinates {(0.95996,0.7212) (0.97427,0.7266) (0.98823,0.7320) (0.99358,0.7285) (0.99594,0.7326) (0.99718,0.7308)};
 \addplot[line width=1.8pt, mark=*, mark size=3, color=blue, dashed, mark=o, mark options={solid, draw=black}] coordinates {(0.94055,0.7115) (0.94357,0.7136) (0.94963,0.7205) (0.95539,0.7153) (0.96077,0.7144) (0.96563,0.7024)};
 \addplot[line width=1.8pt, mark=*, mark size=3, color=purple, mark=square, mark options={solid, draw=black}] coordinates {(0.94426,0.7184) (0.95082,0.7086) (0.96263,0.7164) (0.97229,0.7160) (0.97958,0.7295) (0.98468,0.7149)};
 \addplot[line width=1.8pt, mark=*, mark size=3, color=green, dashed, mark=diamond, mark options={solid, draw=black}] coordinates {(0.94881,0.7016) (0.95893,0.6985) (0.97307,0.6948) (0.98097,0.6955) (0.98625,0.7005) (0.98995,0.6985)};
 \addplot[line width=2pt, dotted, color=magenta] coordinates {(0.94055, 0.7529) (0.99718, 0.7529)};
 \addplot[line width=2pt,  color=cyan] coordinates {(0.94055, 0.5570) (0.99710, 0.5570)};
 \addplot[line width=2pt,  color=brown] coordinates {(0.94055, 0.6589) (0.99718, 0.6589)};
\end{axis}
\end{tikzpicture}
} \hfil
\ref{namedNER}
\caption{NER results relying on sparse coding of different word representations. The x-axis shows the sparsity of the representations with ticks at $\lambda=0.05,0.1,0.2,0.3,0.4,0.5$.}
\label{fig:conllNERresults}
\end{figure*}

We also present in Table~\ref{conllU} the state-of-the-art results of the bidirectional LSTM models by \newcite{plank-sogaard-goldberg:2016:P16-2} for comparative purposes. Note that the authors reported results only on a subset of UD~v1.2 (i.e.~treebanks with at least 60k tokens), for which reason we can include their results on 21 treebanks. Out of these 21 UD~v1.2 treebanks there are 15 and 20 cases, respectively, for which SC and WI+SC produces better results than bi-LSTM\textsubscript{w}. Only FR\textsubscript{w+c} and bi-LSTM\textsubscript{w+c} -- models which enjoy the additional benefit of employing character-level features besides word-level ones -- are capable of outperforming SC and WI+SC.

\subsection{Named entity recognition experiments}

Besides the POS tagging experiments, we investigated if the very same features as the ones applied for POS tagging can be utilized in a different sequence labeling task, namely named entity recognition. In order to evaluate our approach, we obtained the English, Spanish and Dutch datasets from the 2002 and 2003 CoNLL shared tasks on multilingual named entity recognition \cite{TjongKimSang:2002:ICS:1118853.1118877,TjongKimSang:2003:ICS:1119176.1119195}.

We use the train-test splits provided by the organizers and report our NER results using the F1 scores based on the official evaluation script of the CoNLL shared task. Similar to \newcite{Collobert:2011:NLP:1953048.2078186} we also apply the 17-tag IOBES tagging scheme during training and inference. The best F1 scores reported for English by \newcite{Collobert:2011:NLP:1953048.2078186} without employing additional unlabeled texts to enhance their language model is 81.47. When pre-training their neural language model on large amounts of Wikipedia texts they report an F1 score of 87.58.

\begin{table}
\centering
\small
\subfloat[Sparse ($m=1024, \lambda=0.1$)]{
\begin{tabular}{p{2cm}|ccc|c}
         &   en  &   es  &   nl  & Avg.\\\hline
\texttt{polyglot}\textsubscript{SC} & 82.92 & 77.03 & 72.66 & 77.54 \\
CBOW\textsubscript{SC}     & 83.40 & 75.51 & 71.36 & 76.76 \\
SG\textsubscript{SC}       & 82.83 & 75.22 & 70.86 & 76.30 \\
Glove\textsubscript{SC}    & 82.31 & 75.78 & 69.85 & 75.98 \\
\end{tabular}}\\
\subfloat[Dense]{
\begin{tabular}{p{2cm}|ccc|c}
         &   en  &   es  &   nl  & Avg.\\\hline
\texttt{polyglot} & 78.80 & 70.13 & 65.58 & 71.50 \\
CBOW     & 72.68 & 64.49 & 64.80 & 67.32 \\
SG       & 74.68 & 66.17 & 63.95 & 68.27 \\
Glove    & 74.33 & 65.11 & 57.73 & 65.72 \\
\end{tabular}}
\caption{Comparison of the performance of sparse and dense word representations for NER.}
\label{table:sparse_vs_denseNER}
\end{table}

Figure~\ref{fig:conllNERresults} includes our NER results obtained when relying on different word embedding representations as input for sparse coding and different levels of sparsity. Similar to our POS tagging experiments, using \texttt{polyglot}\textsubscript{SC} vectors tend to perform best for NER also. A substantial difference compared to the POS tagging results, however, is that NER performances do not degrade for extreme levels of sparsity neither and that sparse coding-based models perform much better when compared to the stronger FR\textsubscript{w+c} baseline.

In Table~\ref{table:sparse_vs_denseNER}, we compare the effectiveness of models relying on sparse and dense word representations for NER. In order not to fine-tune hyperparameters for a particular experiment, similarly to our previous choices $m$ and $\lambda$ are set to $1024$ and $0.1$, respectively. Results in Table~\ref{table:sparse_vs_denseNER} are in line with those reported in Table~\ref{table:sparse_vs_dense} for POS tagging.

\section{Conclusion}
In this paper we showed that it is possible to train sequence models that perform (near) state-of-the-art on a variety of languages for both POS tagging and NER. Our approach does not require word identity features to perform reliably, furthermore, it is capable of achieving comparable results to traditional feature-rich models. We also illustrated the advantageous generalization property of our model as it retained 89.8\% of its original average POS tagging accuracy when trained only on 1.2\% of the total accessible training sentences.

As \newcite{DBLP:journals/corr/MikolovLS13} pointed out the similarities of continuous word embeddings across languages, we think that our proposed model could be employed in not just multi-lingual, but cross-lingual language analysis settings. In fact, we consider investigating its feasibility as our future work. Finally, we make the sparse coded word embedding vectors publicly available in order to facilitate the reproducibility of our results and to foster multilingual and cross-lingual research.

\section*{Acknowledgement}
The author would like to thank the TACL editors and the anonymous reviewers for their valuable feedbacks and suggestions.

\bibliographystyle{acl2012}
\bibliography{pos_bibs}

\end{document}